\documentclass[conference]{IEEEtran}
\usepackage[left=0.64in,right=0.64in,top=0.75in,bottom=1in]{geometry}
 \makeatletter
 \def\ps@headings{%
 \def\@oddhead{\mbox{}\scriptsize\rightmark \hfil \thepage}%
 \def\@evenhead{\scriptsize\thepage \hfil \leftmark\mbox{}}%
 \def\@oddfoot{}%
 \def\@evenfoot{}}
 \makeatother
\usepackage{amsmath}
\usepackage{algorithm}
\usepackage{url}
\usepackage{graphicx}
\usepackage{multirow}
\usepackage{multicol}
\usepackage{verbatim}
\usepackage{todonotes}
\usepackage{amsfonts}
\usepackage{indentfirst}
\usepackage[noend]{algpseudocode}
\usepackage{caption}
\usepackage{subcaption}
\usepackage{array}
\usepackage{bm}
\usepackage{color}
\usepackage{verbatim}
\usepackage{enumerate}
\usepackage[algo2e]{algorithm2e}

\DeclareMathOperator*{\argmin}{arg\,min}

\usepackage{xspace}
\usepackage{cite}

\begin{document}

\title{Implicit Context-aware Learning and Discovery for Streaming Data Analytics}

\author{\IEEEauthorblockN{Kin Gwn Lore, Kishore K. Reddy}
\IEEEauthorblockA{United Technologies Research Center, East Hartford, Connecticut 06118 \\
 Email:\{lorek,reddykk\}@utrc.utc.com}
}

\maketitle

\begin{abstract}
The performance of machine learning model can be further improved if contextual cues are provided as input along with base features that are directly related to an inference task. In offline learning, one can inspect historical training data to identify contextual clusters either through feature clustering, or hand-crafting additional features to describe a context. While offline training enjoys the privilege of learning reliable models based on already-defined contextual features, online training for streaming data may be more challenging--- the data is streamed through time, and the underlying context during a data generation process may change. Furthermore, the problem is exacerbated when the number of possible context is not known. In this study, we propose an online-learning algorithm involving the use of a neural network-based autoencoder to identify contextual changes during training, then compares the currently-inferred context to a knowledge base of learned contexts as training advances. Results show that classifier-training benefits from the automatically discovered contexts which demonstrates quicker learning convergence during contextual changes compared to current methods.
\end{abstract}



\section{Introduction}\label{sec:intro}

Contextual cues can greatly benefit learning of predictive tasks in a machine learning model. A single datapoint may be meaningless. However, if the scope is expanded to include the context from which the data is obtained, one can obtain greater insights and perhaps substantially improve decision making. Context can exist in the form of semantic context, spatial context, and scale context in the domain of computer vision. These contexts, when leveraged, has shown great success in various works from object detection to image segmentation~\cite{wang2017attribute,japuria2017casnsc,li2016object,liu2017face}. Another application area concerns time-series data obtained from sensors or transactional in nature, where contextual features could be derived from time, weather, or event logs~\cite{matei2017context,gao2017detecting,liu2016multi}. Furthermore, mobile devices can use contextual features such as location, infer user activities based on on-board accelerometers, and user events to provide tailored mobile services or improving user experience~\cite{mezhoudi2013user,kabir2015machine,wang2010camf,zhou2012context,kwapisz2011activity}. However, such contextual features are often \textit{explicit} in nature as they are specially selected and incorporated to fit the purpose of the machine learning task. However, we are interested in automatically identifying context changes without knowing what are the contexts.

\begin{figure}[htb]
\vspace{-8pt}
    \centering
    \includegraphics[width=0.85\columnwidth,trim={40, 80, 40, 40},clip=True]{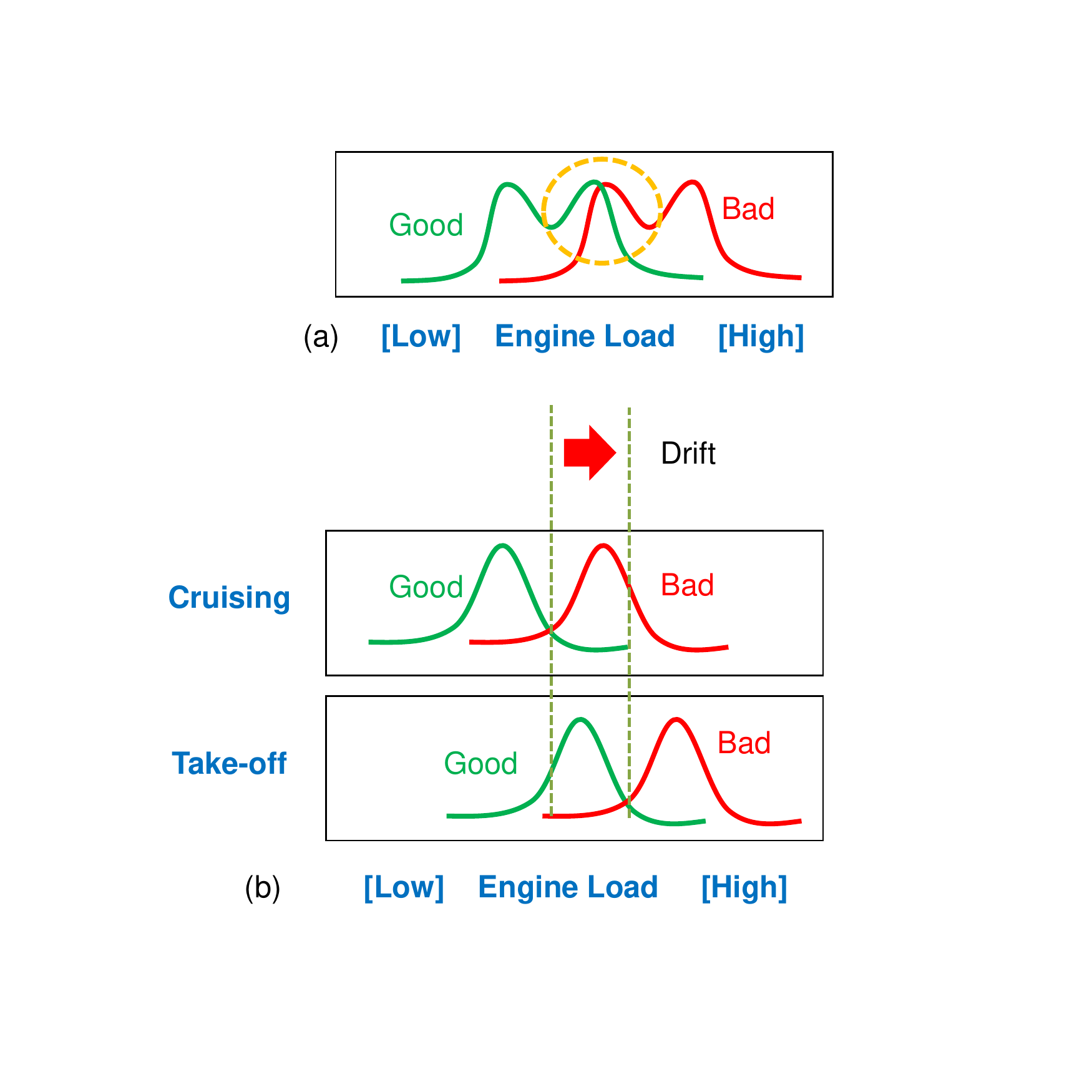}
    \caption{Illustration on the importance of contextual features in a classification task.}
    \label{figure:enginectxt}
     \vspace{-8pt}
\end{figure}

Consider the illustration shown in Fig~\ref{figure:enginectxt}. Suppose there is an engine data with various sensors attached from which we can derive the engine load. Suppose there are two classes present in the data, one being the engine under excessive load (positive event) whereas the other engine is under nominal load (negative event). If context is not known, the data distribution of the two classes can be represented in Fig~\ref{figure:enginectxt}(a). If the model is given a task to discriminate between nominal or excessive loads, it is inevitable that an under-performing model will be learned due to contradictory evidences observed in data within high-confusion regimes. On the other hand, consider contextual drift that happens along a third dimension---in this case, it is time. As time progresses, an aircraft might change its mode of operation from taking-off to cruising. In this case, we do not explicitly identify time as a contextual feature as it has no predictive power. We only know that the data distribution has changed in the passage of time---caused by some changes context---which is \textit{implicit} in nature. Given these contextual clues (Fig.~\ref{figure:enginectxt}(b), there is a clear decision boundary that separates between nominal and excessive engine loads, which in turn provides more useful information for a successful classification model.

Usually, in offline learning, one can inspect historical training data to identify contextual clusters either through feature clustering, or hand-crafting additional features to describe a context. While offline training enjoys the privilege of learning reliable models based on already-defined contextual features, online training for streaming data may be more challenging---the data is streamed through time, and the underlying context during a data generation process may change. Furthermore, the problem is exarcebated when the number of possible context is not known. In this paper, our contributions are threefold:
\begin{enumerate}
    \item A novel online-learning algorithm is proposed;
    \item Contrary to handcrafting contextual features to be used as part of the input, the algorithm discovers new contexts as new samples arrive;
    \item The algorithm is compared with multiple baseline approaches over four different applications and datasets;
\end{enumerate}

This paper is organized as follows. In section~\ref{sec:related}, we outline related works and highlight the contributions of the present work. In section~\ref{sec:problem}, we setup the problem and describe how one might choose to approach the described problem. Then, we present the proposed method and outline the online learning algorithm in section~\ref{sec:method}. Multiple experiments are performed to benchmark the proposed method against the non-context aware counterparts, where the results are presented and discussed in section~\ref{sec:results}. Finally, we summarize the paper and end with conclusions in section~\ref{sec:conclusions}.

\section{Related works}\label{sec:related}
In the context of concept and context drift, authors of~\cite{widmer1996recognition} proposed the MetaL(B) and MetaL(BI) as meta-learning frameworks to identify potential contextual clues during online training. However, the method is limited to a problem set where external contextual features are already known and prespecified; contextual changes are detected by changes in contextual features. In~\cite{widmer1996learning}, the authors presented an approach to learn in the presence of concept drift and hidden contexts where object descriptions from the past are stored in addition to detecting the changes to the current hypothesis. Descriptions are specified for each positive and negative groups, and the running count of the descriptions are tracked.

On the other hand, there are applications that uses neural network-based approaches for streaming data. However, most applications are directed to anomaly detection where autoencoders are used to detect anomalies in streaming data. Authors of~\cite{dong2018threaded} observed improved anomaly detection by implementing threaded ensembles of autoencoders, while authors of\cite{zhou2012online} features using denoising autoencoders to effectively learn features during online learning.

Further, there are existing works which combines both worlds on learning under context/concept drifts with neural network-based approaches. \cite{yan2016correcting} highlights the use of autoencoders to learn a drift-corrected representation of the dataset for the purpose of transfer learning. Additionally, some authors decided to handle concept drift in online streaming data by constructing a new classifier for every new encountered context and contexts~\cite{budiman2016adaptive}.

Recently, neural networks has gained immense traction in the advent of deep learning with the rise of computing power. Autoencoders, being a class of neural networks, has been known to perform particular well in identifying hidden representations in data via an unsupervised learning method. This provided a huge advantage where clusters of feature representations can be learned without the need of explicit labels by experts. The algorithm proposed in the present work leverages the power of autoencoders to identify underlying contexts, which then relate the current learning task with the inferred context. More details follow in the next section. To the best of our knowledge, at this point there are no existing work that identifies new context on-the-fly and dynamically appends the contextual feature to the data matrix without handcrafting contextual features.

\section{Problem setup}\label{sec:problem}

Consider an online learning task centered around classification. Data, or the feature matrix, arrives as a $d$-dimensional vector $x:\mathcal{R}^d$ associated with the class labels $y\in \{0,1\}$. At every time step $t$, the model receives a new data sample to train the classification model $f(\cdot)$. At each newly-arriving samples, the model will perform class predictions and compare the predicted output $\hat{y}$ with the supplied ground-truth label $y$. Any wrongly-classified model will provide feedback to the model to update its learned parameters.

A \textit{concept shift} happens whenever there is a change in the relationship between supplied features $\mathcal{X}$ and the corresponding target values $\mathcal{Y}$. Typically, this happens whenever there is a change in the process which results from a change in the mode of operation. Consider the operation of a jet engine; suppose in one mode of operation, monitoring specific feature values observed in attached sensors can provide insights to the health of the engine. However, when a change occurs such that the engine is operating under a different mode, what used to be healthy could now be considered as unhealthy (i.e., where the decision threshold has changed which affects the labeling strategy). In a such a scenario, typical online-learning algorithms which do not incorporate context awareness will fail to adjust to this change in context.

To evaluate the proposed algorithm, we present three datasets that can simulate the aforementioned problem setup. We first have a look at the \texttt{stagger} dataset~\cite{widmer1993effective}, which comprised of basic attributes such as shapes, sizes, and color, and the model shall predict a binary value where the labeling logic is hidden behind the training data that switches at fixed intervals. Next, we demonstrate on the \texttt{MNIST-digits} dataset~\cite{lecun1998mnist}, where the task is to perform digit classification while the two datasets switch back-and-forth during training at fixed intervals. Finally, we evaluate the algorithm on the Naval Propulsion dataset from the UCI Data Repository (\texttt{propulsion})~\cite{coraddu2016machine} in the context of condition-based maintenance. Each of these datasets are briefly described below.

\subsection{The \texttt{stagger} dataset}
The \texttt{stagger} dataset~\cite{widmer1993effective} is defined in a simple block-world defined by three nominal attributes:
\begin{itemize}
    \item size $\in$ \{small, medium, large\}
    \item color $\in$ \{red, green, blue\}
    \item shape $\in$ \{square, circular, triangular\}
\end{itemize}

Additionally, three target concepts are defined and enumerated below:
\begin{enumerate}
    \item size = small $\land$ color = red
    \item color = green $\vee$ shape = circular
    \item size = (medium $\vee$ large)
\end{enumerate}

\begin{figure}[htb]
\vspace{-8pt}
    \centering
    \includegraphics[width=0.7\columnwidth,trim={10 10 10 0},clip=True]{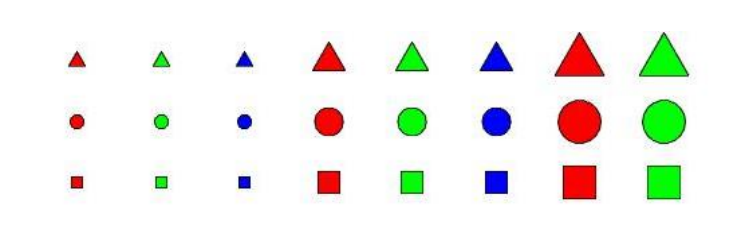}
    \caption{The \texttt{stagger} dataset is defined by objects with three nominal attributes: size, color, and shape, each with three values per attribute.}
    \label{figure:staggerdataset}
     \vspace{-8pt}
\end{figure}

The hidden target concept will switch from one of these definitions to the next, following the sequence 1-2-3-1-2-3, which results in extreme concept drift. The key metric to evaluate the performance of the model then hinges on the ability to converge to high accuracy at a rate quicker than the baseline approaches. In the experiments, each time step consists of a sample, and the concept switches every 200 samples. Since there are a total of 6 concepts (3 concepts, each repeated twice), there are a total of 1,200 samples available for online training and evaluation.

\subsection{The \texttt{MNIST-digits} dataset}
The \texttt{MNIST} dataset~\cite{lecun1998mnist} is a well-known benchmarking dataset for digits recognition (digits of 0-9) with pixel dimensions of $28\times 28$-px and grayscale in nature. Like \texttt{MNIST}, the \texttt{digits} dataset is also a database of handwritten digits obtained from \texttt{scikit-learn}, originally based on the USPS dataset~\cite{seewald2005digits} with image dimensions of $8 \times 8$-px in a vectorized format (i.e., 64-dimensional). To bring both datasets onto a common space and keep computational complexity low, we resized the \texttt{MNIST} digits into the common $8 \times 8$-px dimension as the \texttt{digits} dataset.

\begin{figure}[htb]
\vspace{-8pt}
    \centering
    \includegraphics[width=\columnwidth,trim={10 10 10 10},clip=True]{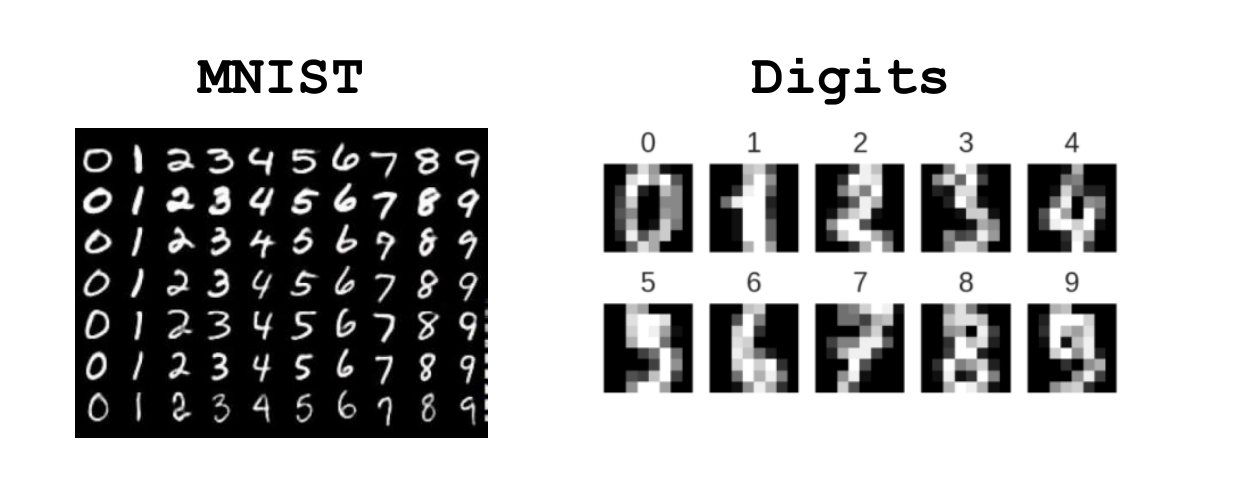}
    \caption{The \texttt{MNIST} dataset is a collection of digits ranging from 0-9.}
    \label{figure:mnistdataset}
     \vspace{-8pt}
\end{figure}

Each sample will come from a fixed dataset (either \texttt{MNIST} or \texttt{digits}) which lasts for 1,000 samples. We simulate training data arrival by randomly sampling from the superset which contains 50,000 samples from \texttt{MNIST} and 1,797 samples from \texttt{digits}. Denoting \texttt{MNIST} as $M$ and \texttt{digits} as $D$, the context-switching sequence is M-D-M-D-M-D which totals 6,000 samples.

\subsection{The \texttt{propulsion} dataset}
The UCI Repository Naval Propulsion Plants dataset~\cite{coraddu2016machine} is a real-world dataset collected for the development of machine-learning driven condition-based maintenance models. The dataset contains sensor readings and parameters such as \textit{lever position}, \textit{ship speed}, \textit{gas turbine shaft torque}, \textit{propeller torque}, etc. along with real-valued information such as \textit{compressor decay} and \textit{turbine decay}. To frame it as a condition-based maintenance model, we are interested in determining whether the compressor state is healthy or unhealthy (i.e., a binary classification problem). The dataset has 16 features (thus 16-dimensional) so the interactions between features is moderately high-dimensional. Based on the statistics of the dataset, we determine that the compressor is in a healthy state if the responding real-valued variable \textit{compressor decay} is higher than percentile $q(\cdot)$ derived from the data. With this, the switching contexts are:
\begin{enumerate}
    \item Label = Unhealthy (1) \textbf{if} $q$(\textit{Compressor decay})$>0.1$
    \item Label = Unhealthy (1) \textbf{if} $q$(\textit{Compressor decay})$>0.9$
\end{enumerate}
and Label is Healthy (0) if the conditions are not met. Each operating modes last for 300 samples and the sequence are defined as C1-C2-C1-C2 (totalling 1200 samples).

\section{Proposed method}\label{sec:method}

\begin{figure}[htb]
    \centering
    \includegraphics[width=0.8\columnwidth,trim={20 20 0 30},clip=True]{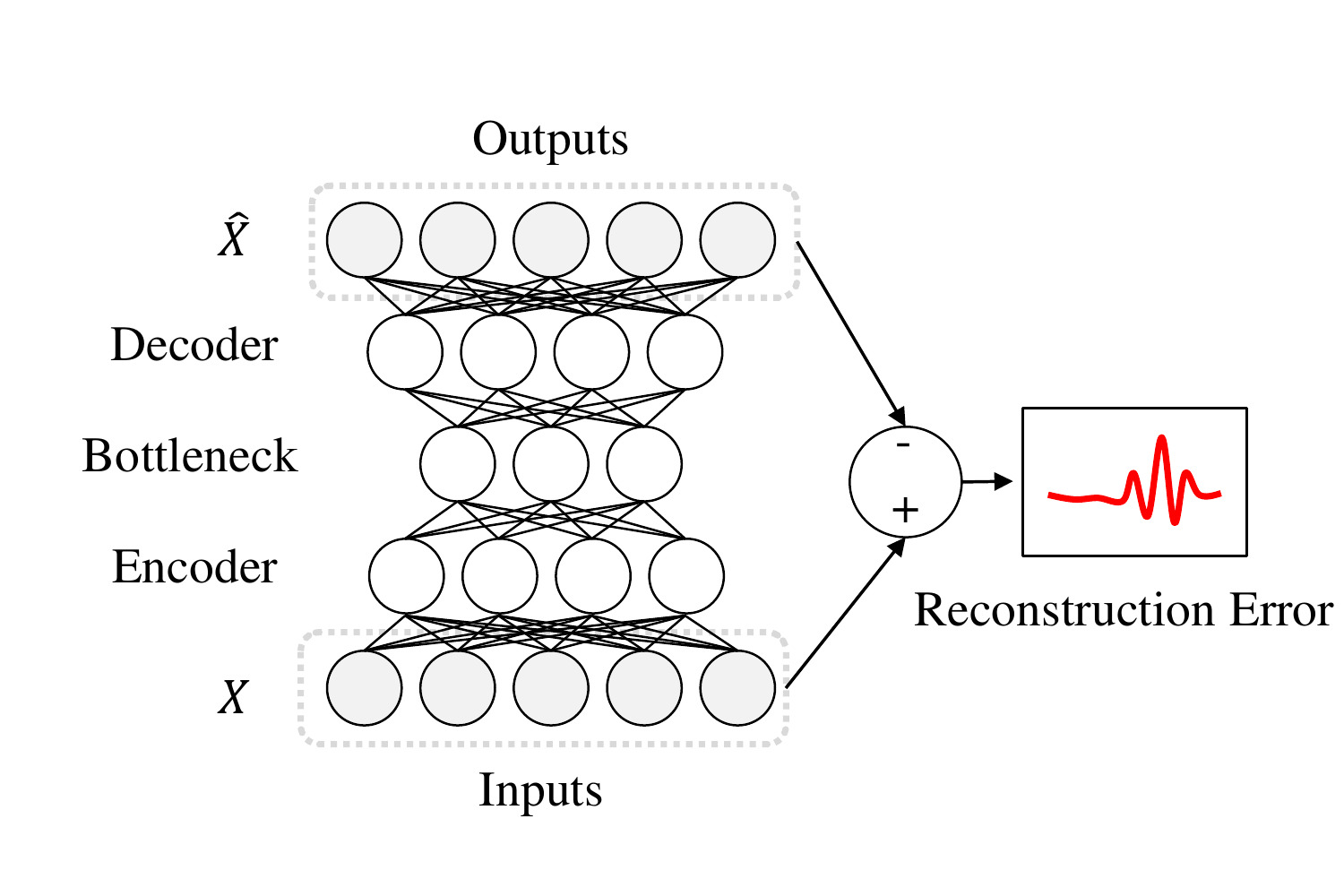}
    \caption{A neural network-based autoencoders learn latent representation in the underlying normal data, where the code vector is encoded in the bottleneck layer. The DAE attempts to reconstruct the input based on this compressed representation of the input. The reconstruction error between the input and the output is used directly for anomaly detection.}
    \label{figure:sae}
\end{figure}

\textbf{Prerequisites: autoencoders.}
The autoencoder $g(\cdot)$~\cite{vincent2010stacked,hinton2006reducing} (fig. \ref{figure:sae}) is able to automatically learn useful features from the input data $X$ and reconstruct the input data based on the learned features. It showed promising results across a wide variety of cross-domain anomaly detection problems~\cite{sakurada2014anomaly,martinelli2004electric,nolle2016unsupervised}. The autoencoder is trained to learn a low-dimensional representation of the normal input data $X$ by attempting to reconstruct its inputs to get $\hat{X}$ with the following objective function:
\[ \theta = \argmin_\theta \mathcal{L}(X,g(X)) \]
where $\theta$ is the parameters of the autoencoder $g$ (i.e., weights of the neural network), $\mathcal{L}$ is the loss function (typically $\mathcal{L}_2$ loss), $X$ is the input, and $\hat{X}=g(X)$ is the reconstruction of the autoencoder. When presented with data that comes from a different data-generating distribution, it is expected that a high reconstruction error to be observed. The errors can be modeled as a normal distribution---an anomaly (in this case, out-of-context data) is detected when the probability density of the average reconstruction error is below a certain threshold.

\textbf{Proposed method.} In this work, we propose a novel method for context-aware incremental learning by jointly training a neural network-based autoencoder $g$ for contextual inference and a classifier for the designated task. In general, there is only a single classifier $f(\cdot)$ driven by the classification task. However, during the learning phase, the autoencoder receives features $x$ and the data labels $y$ to learn the underlying representation of the features and the labels $z = x \oplus y$. Through this approach, a nominal \textit{concept representation} for a particular operating mode is derived. For this approach to function, we place the assumption where classification accuracy will experience a sharp decrease in the event of concept drift. Hence, a sharp decrease of classification accuracy indicates a potential change in concept, which triggers the autoencoder $g_i$ to compare the representation of the new data with the average representation of the learned context computed via $g_i$. If the representation is substantially different (i.e., high reconstruction error), a new concept is hypothesized, where then a \textit{contextual variable} $c$ is introduced as a flag for the new concept. The subsequent training data matrix will be denoted as $x\leftarrow x\oplus c$ which indicates that the newly added contextual variable is now part of the feature matrix as inputs to the classifier.

\begin{figure*}[htb]
\vspace{-8pt}
    \centering
    \includegraphics[width=0.75\textwidth,trim={10 40 10 20},clip=True]{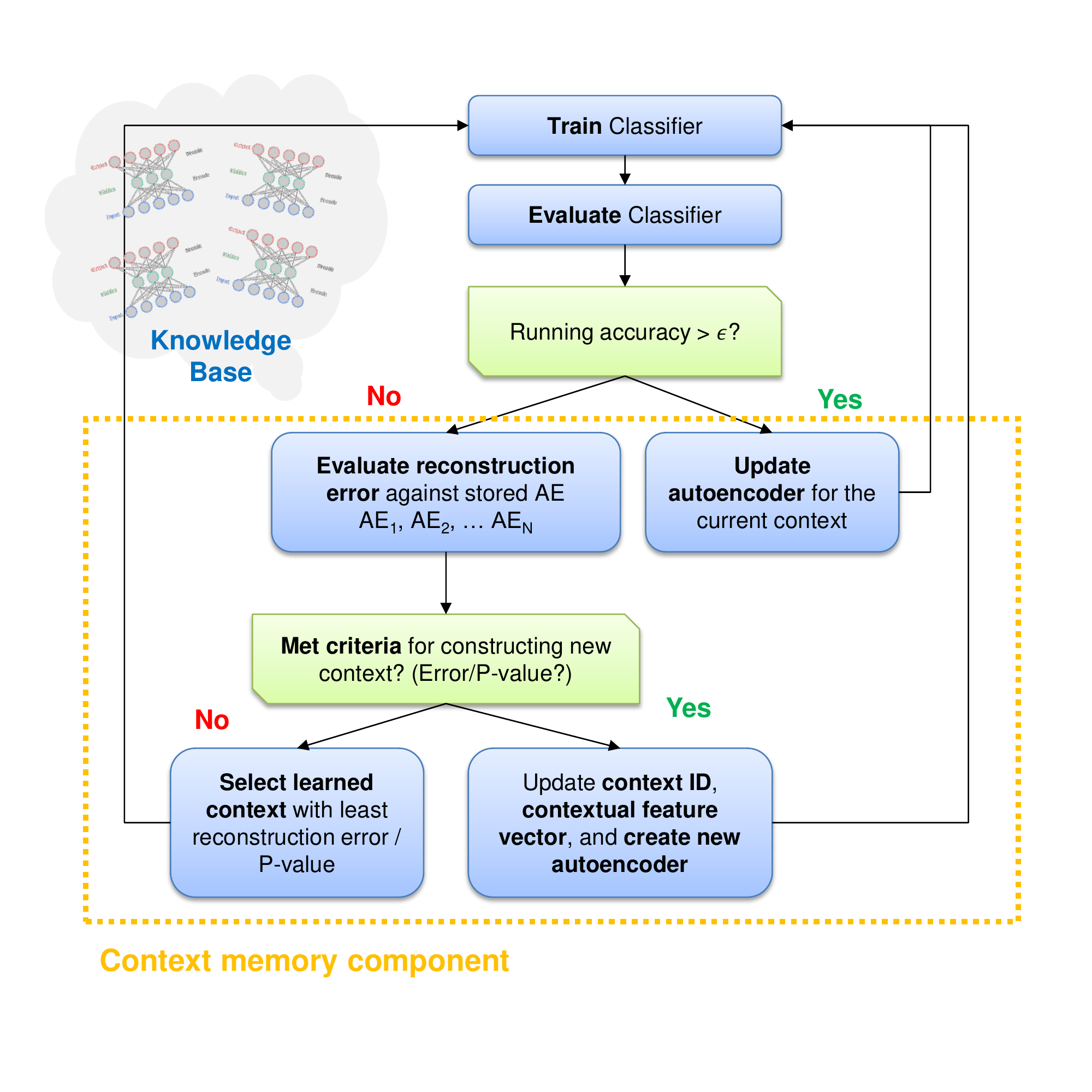}
    \caption{Flowchart of the proposed algorithm.}
    \label{figure:frameworkflowchart}
\end{figure*}

One autoencoder is learned for each concept as a descriptor for the training instance. During every encounter of potential concept switch, the new sample is evaluated against a knowledge base of autoencoders $\mathcal{G}=\{ g_1,g_2,g_3,...,g_{n_c} \}$ to derive the reconstruction errors where $n_c$ is the number of seen (and hypothesized) contexts. If for all autoencoders the reconstruction error $\epsilon = \mathcal{L}(z,\hat{z})$ is above a certain threshold, then a new context is hypothesized. On the other hand, if one of the autoencoder returns a reconstruction error falling within a predefined threshold (based on statistical significance), then the new sample is thought to come from a context that has been encountered previously. Using this framework, the online learning algorithm is able to adapt to changing concepts and contexts without requiring prior complete knowledge on the number of possible concepts present in the data. The framework is summarized in Fig.~\ref{figure:frameworkflowchart}.

In the framework, we modeled reconstruction errors as a normal distribution. Under a fixed context or concept, the mean and standard errors of the reconstruction error specific to a context is monitored. As previously mentioned, whenever the online training experiences a sharp drop in training accuracy, we pass the current training sample $z=x\oplus y$ through the autoencoder $g_i$ related to the current context to compute its reconstruction error $\epsilon = \mathcal{L}(z,\hat{z})$. Given a past history of verified samples coming from the evaluated context, we can derive $\mu_\epsilon$ and $\sigma_\epsilon$. For a new sample under decreased accuracy, we consider a sample as out-of-context if:
\[ P\left(\frac{\epsilon - \mu_\epsilon}{\sigma_\epsilon}\right) < k \]
where $k$ is a defined threshold. The online training algorithm is outlined in algorithm~\ref{alg:algo}.

\begin{algorithm}[H]
\SetAlgoLined
\textbf{Initialization:}\\
- Initialize current context ID, $n_c \leftarrow 1$\\
- Initialize historical accuracies, $acc \leftarrow \phi$\\
- Initialize classifier, $f(\cdot)$\\
- Initialize context autoencoder, $g_1(\cdot)$\\
- Initialize autoencoder list,  $\mathcal{G} \leftarrow g_1$\\
- Define accuracy threshold, $t\leftarrow 0.9$\\
- Define autoencoder window, $T\leftarrow 20$\\
- Train classifier $f(x_0:x_{t-1},y_0:y_{t-1})$\\
- Train autoencoder $g(x_0 \oplus y_0 :x_{t-1}\oplus y_{t-1}, x_0 \oplus y_0 :x_{t-1}\oplus y_{t-1})$\\
 \While{Data $x_t$ is streaming}{
  Make prediction, $\hat{y} \leftarrow f(x_t)$\\
  Evaluate accuracy, $acc_i \leftarrow I\left[\hat{y},{y}\right]$\\
  Append accuracy to list, $acc \leftarrow acc \bigcup acc_i$\\\
  \eIf{mean($acc_{t-T}:acc_{t}$) $>t$}{
   Select $\chi \leftarrow (x_0\oplus y_0:x_t \oplus y_t | context = n_c)$\\
   Update autoencoder $g_{n_c} \leftarrow g(\chi,\chi)$
   }{
   Pass $x_t\oplus y_t$ through stored list of autoencoders $g_0, g_1, ..., g_{n_c}$\\
   Compute $\epsilon_i$ for each $g_i$ for $i\in (0,1,..., n_c) $\\
   \eIf{$P\left(\frac{\epsilon_i - \mu_{\epsilon_{g_i}}}{\sigma_{\epsilon_{g_i}}}\right) < k$}
       {
       Update autoencoder list $G\left[n_c\right] \leftarrow g_{n_c}$\\
       Increment context ID $n_c \leftarrow n_c+1$\\
       Initialize new autoencoder $g_{n_c}$
       }{Assign context ID $n_c \leftarrow \argmin{\epsilon_i}$
       }
  }
 }
 \caption{Implicit Context-aware Learning on Streaming Data}
 \label{alg:algo}
\end{algorithm}

One apparent advantage of this framework is that for each new concepts encountered, the current concept is stored as a \textit{contextual variable} to aid decision support. While not directly linked to the class labels, contextual variables are important where the decisions can be conditioned upon the contextual variables. One problem arises when during online predictions (without training data), there is no way to know the true labels. This prevents the autoencoder as it requires both the data and label to reconstruct from learned representation. From here, there are two possible ways to go forward: (1) One can define a placeholder default value in place of the label, and (2) an expert can inspect the data and provide the contextual flag explicitly by examining past contexts. In (1), using default values in place of the expected labels will not provide a reliable estimate of reconstruction error, but by waiting a few more time steps to gather a batch of data to evaluate against the autoencoders to obtain an average representation, we will be able to compare the representations via the autoencoder with higher certainty. On the other hand, method (2) can first trigger an alarm to grab the attention of the expert, where the expert can use human reasoning to assign a context, after which the context is defined for all subsequent data.

\section{Results and discussions}\label{sec:results}

\subsection{Compared approaches}
In our method, we emphasize on \textit{context awareness}, where for every new concepts encountered, a new contextual variable is defined as the input feature. We compared the online training framework against baseline approaches without context awareness, that is, the context is assumed to be the same without constructing new contextual features. For clarity, we designate the following names and description for the algorithms:
\begin{enumerate}
    \item \textit{*Implicit Context-aware Learning with Memory component (ICAL-Mem)}: The proposed approach with context awareness achieved through autoencoders. The classifier is a tree-based model due to its ease of training.
    \item \textit{*Implicit Context-aware Learning (ICAL)}: A simplified version of the proposed approach, but without context awareness. This approach does not store autoencoders as concept descriptors; rather, a drop in online classification accuracy directly signifies a change in context and all information related to previously learned concepts are forgotten.
    \item \textit{Non-context-aware Learning (Non-CAL)}: A classifier which does not respond to contextual changes and learns from a long history of available data, at a time scale larger than the time span of a particular context. A CART algorithm is selected for this approach.
    \item \textit{Myopic Non-context-aware Learning (Myopic Non-CAL)}: A classifier that learns from data coming from a short considered sliding window. Such approach is expected to respond to misclassifications as old knowledge is purged and forgotten from training. A CART algorithm is also selected for this approach.
\end{enumerate}
Methods with an asterisk (*) are methods proposed in this paper which is centered upon context-aware learning.

\subsection{Evaluation metric}
During online training, the predicted label and the true label of a new sample is stored. A running exponentially-weighted moving average (EWMA) of the classification accuracy is presented, and then the average accuracy for the whole experiment is computed. If an algorithm takes a slow recovery of classification accuracy whenever a new context is encountered, the resultant post-experiment average accuracy will be lower than the algorithm that undergoes fast recovery. The metric is intuitive, and is the usual arithmetic mean of accuracy defined as:
\[ \text{Accuracy} = \frac{1}{K}\sum_{i=0}^{K} I(\hat{y}_i,y_i) \]
where $K$ is the number of evaluation time steps, $I(\cdot)$ is the evaluation function (e.g. rolling mean of classification accuracy) operating on the predicted label $\hat{y}$ and the true label $y$ at step $i$.

\subsection{Results on \texttt{stagger} dataset}
\begin{figure}[htb]
\vspace{-8pt}
    \centering
    \includegraphics[width=\columnwidth,trim={0 0 0 0},clip=True]{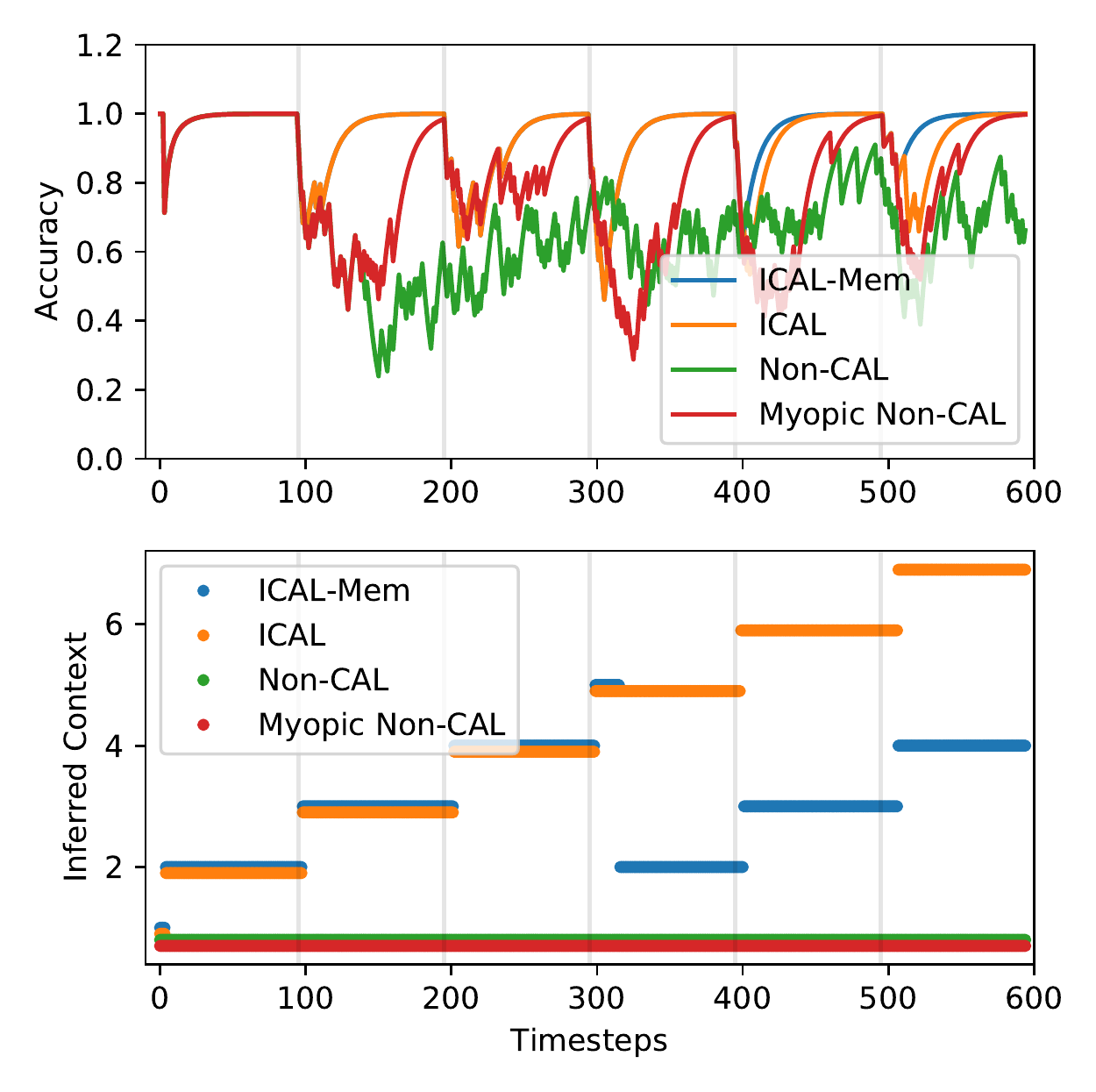}
    \caption{(Top) EWMA accuracies of all considered algorithms evaluated on the \texttt{stagger} dataset. Concept switch is indicated by the grey vertical lines; the 6 partitions corresponds to concepts 1-2-3-1-2-3. (Bottom) Inferred contexts during online training.}
    \label{figure:staggerresults}
\end{figure}

The top plot in Fig.~\ref{figure:staggerresults} shows the EWMA accuracies of the considered algorithms evaluated on the \texttt{stagger} dataset. During the first context, all algorithms performed equally well and converges to good accuracies at the same rate. As the data advances into a new context, all algorithms suffered a significant drop in accuracy, but the awareness to changing contexts enabled ICAL-Mem and ICAL to recover quickly. Non-CAL and Myopic Non-CAL on the other hand, without contextual variables, struggles to identify underlying patterns due to conflicting evidence during previous learning. However, myopic Non-CAL was able to recover quicker than Non-CAL by forgetting older contradictory evidences from the previous contexts.

The bottom plot in Fig.~\ref{figure:staggerresults} shows the identified context using the autoencoder approach. Being aware to contextual changes, ICAL constantly increments the number of contexts by 1 and starts relearning without considering old memories. ICAL-Mem is able to do better by \textit{remapping} the new incoming sample to a \textit{previously-learned context}, and subsequently switches training mode to condition upon a previously-learned context. Hence, it was able to identify the same contexts during every second encounter. Note that during the second encounters (i.e., partitions 4, 5, 6) the EWMA accuracies is much higher than the ICAL without memory, as well as the non-context aware counterparts (i.e., Non-CAL and Myopic Non-CAL).

\subsection{Results on \texttt{MNIST-digits} dataset}

Results for this dataset is presented in the same format as before. For this dataset, the contexts are actually fairly similar which prevents the autoencoder from inferring contexts effectively. Due to the similarity of the \texttt{MNIST} and \texttt{digits} datasets in the feature space, the classification model was still able to perform well despite the context change. One striking difference between this dataset and the \texttt{stagger} dataset is that the feature undergoes some change, not the labeling strategy. Coupled with similar features (i.e., grayscale handwritten digits), there is only a small benefit in using context-aware approaches. This is shown by the Non-CAL method tracing closely with the context-aware counterparts (ICAL-Mem and ICAL). In terms of the inferred context, the first sharp contextual change is idenfified after the first partition. Interestingly, the model has seen enough training data and perform equally well during the second context change. Hence, the classification accuracy did not suffer and consequently not trigger context reidentification.

\begin{figure}[htb]
\vspace{-8pt}
    \centering
    \includegraphics[width=\columnwidth,trim={0 0 0 0},clip=True]{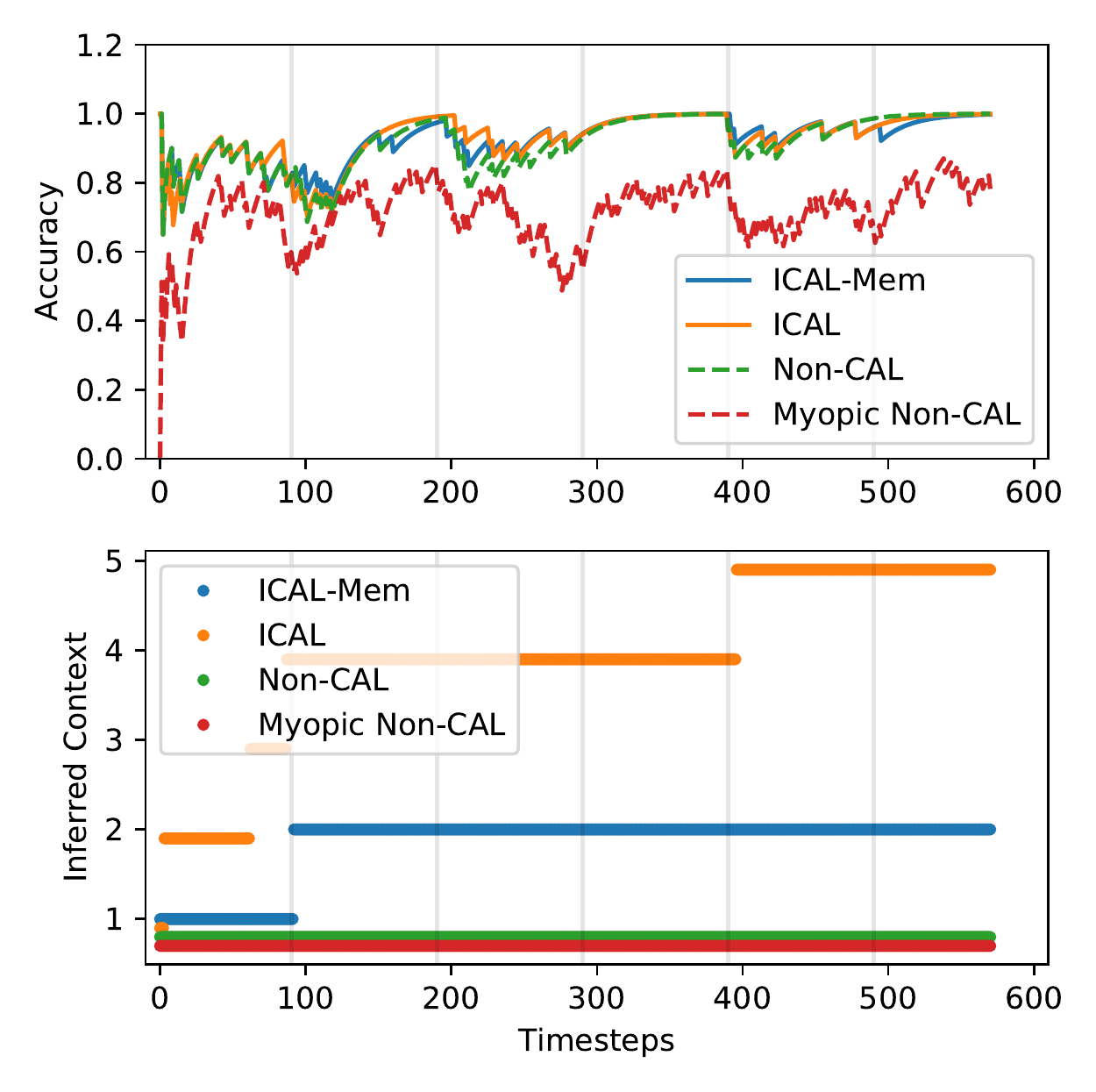}
    \caption{(Top) EWMA accuracies of all considered algorithms evaluated on the \texttt{MNIST-digits} dataset. Context switch is indicated by the grey vertical lines; the 6 partitions corresponds to concepts 1-2-1-2-1-2. (Bottom) Inferred contexts during online training.}
    \label{figure:results_digits}
\end{figure}

\subsection{Results on \texttt{Propulsion} dataset}

\begin{figure}[htb]
\vspace{-8pt}
    \centering
    \includegraphics[width=\columnwidth,trim={0 0 0 0},clip=True]{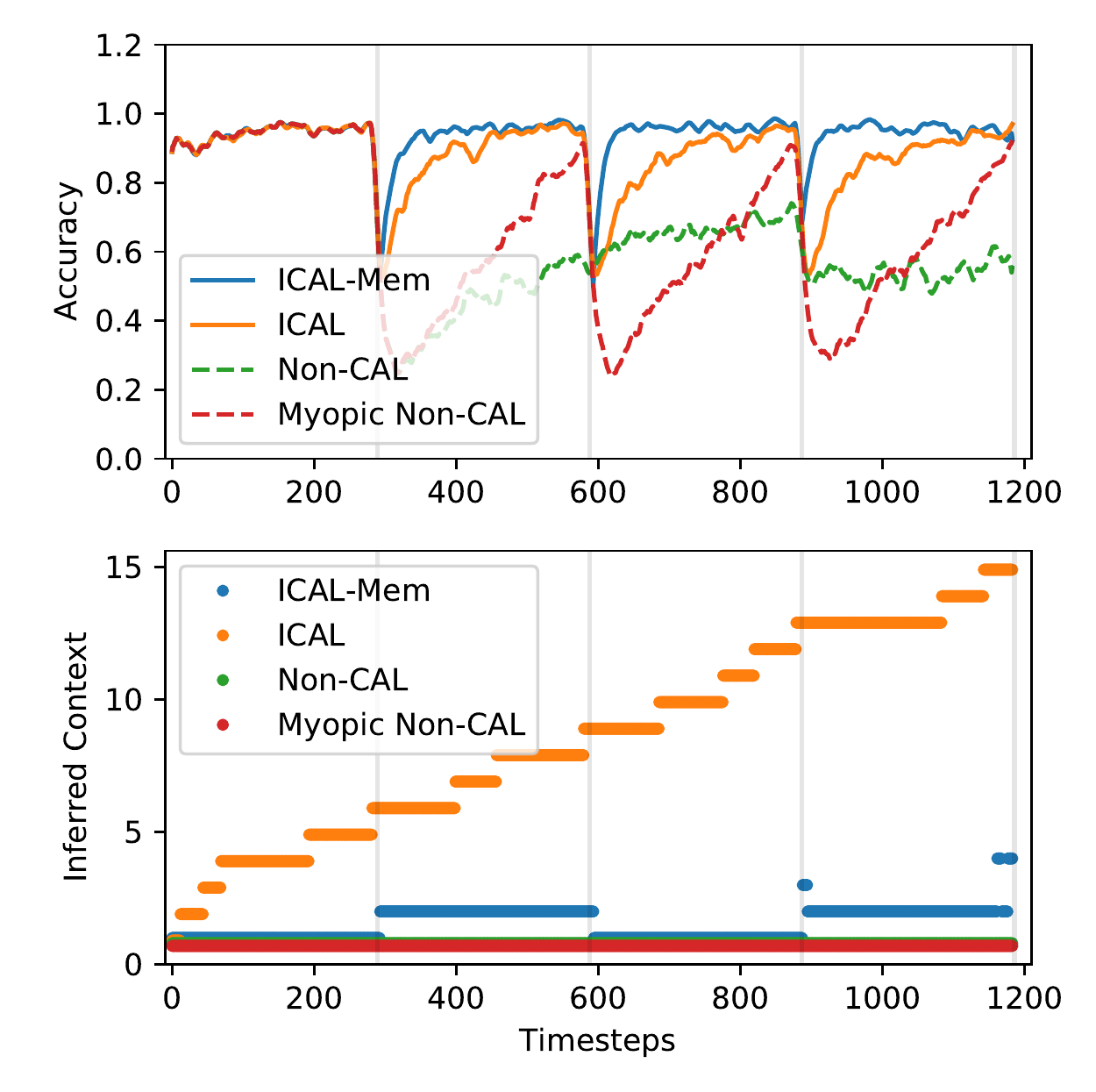}
    \caption{(Top) EWMA accuracies of all considered algorithms evaluated on the \texttt{Propulsion} dataset. Context switch is indicated by the grey vertical lines; the 4 partitions corresponds to concepts 1-2-1-2. (Bottom) Inferred contexts during online training.}
    \label{figure:results_propulsion}
\end{figure}

Results for this dataset is striking---there is a huge gain of implicit context-aware learning with the autoencoder-based memory component. The ICAL-Mem approach clearly recovers faster during every conceptual drift, and correctly maps the inferred context to one that has been learned. As a result of the reassignment of previously-known context, the online classification accuracy quickly converges at a much higher rate compared to other baseline approaches. Myopic Non-CAL requires a long time to recover, while Non-CAL approach was not able to recover quickly due to contrasting evidence in previously learned data. The online accuracy plots are presented in Fig.~\ref{figure:results_propulsion}.

\subsection{Discussions}

Table~\ref{tab:results} shows a summary on the average accuracy throughout the entirety of the online training process. As a reminder, higher average accuracy is an indicator of quick convergence/recovery during a context/concept switch. In general, there is a much greater benefit in incorporating context-awareness during training especially a new context/concept is discovered. The benefit is more apparent whenever there is change in the ground truth compared to changes in features. This is because given enough training data, the model might learn to generalize better to accommodate shifts within the feature space, while it is not possible for the model to generalize to a different label with the similar set of features.

\begin{table}[htb!]
\caption{Averaged accuracies through the entirety of the training process. Methods marked with asterisks (*) are methods proposed in this paper.}
\label{tab:results}
\begin{center}
\scriptsize
\begin{tabular}{|c|c|c|c|c}
\hline	
\textbf{Method}	&	\textbf{\texttt{Stagger}}  & \textbf{\texttt{MNIST-Digits}} & \textbf{\texttt{Propulsion}}	\\\hline
\textbf{*ICAL-Mem}	    & \textbf{93.94}\% &	93.17\% & \textbf{93.77}\%  \\
\textbf{*ICAL}	        & 92.61\% &	\textbf{93.52}\% & 88.11\% \\
\textbf{Non-CAL}	    & 68.23\% &  92.29\% & 64.20\% \\
\textbf{Myopic Non-CAL}	& 82.61\% &	72.33\% & 66.00\% \\\hline
\end{tabular}
\end{center}
\end{table}

\section{Conclusions}\label{sec:conclusions}
In this paper, we proposed a novel online-learning algorithm that generates new contextual identifier to aid learning of classification tasks under context or concept drifts. The building blocks of the algorithm consist of a task-oriented classifier, in additional to autoencoders that stores low-dimensional representation of the data which can increase in numbers as new contexts or concepts are discovered. The algorithm is evaluated against several baselines without context-awareness, which is a fundamental component to the success of online learning under drift. Results has shown that up to 30 pts. improvement (percent) can be achieved by incorporating context-awareness during online training, as opposed to the non-context aware counterparts.

\bibliographystyle{IEEEtran}
\bibliography{references}

\begin{thebibliography}{10}
\providecommand{\url}[1]{#1}
\csname url@samestyle\endcsname
\providecommand{\newblock}{\relax}
\providecommand{\bibinfo}[2]{#2}
\providecommand{\BIBentrySTDinterwordspacing}{\spaceskip=0pt\relax}
\providecommand{\BIBentryALTinterwordstretchfactor}{4}
\providecommand{\BIBentryALTinterwordspacing}{\spaceskip=\fontdimen2\font plus
\BIBentryALTinterwordstretchfactor\fontdimen3\font minus
  \fontdimen4\font\relax}
\providecommand{\BIBforeignlanguage}[2]{{%
\expandafter\ifx\csname l@#1\endcsname\relax
\typeout{** WARNING: IEEEtran.bst: No hyphenation pattern has been}%
\typeout{** loaded for the language `#1'. Using the pattern for}%
\typeout{** the default language instead.}%
\else
\language=\csname l@#1\endcsname
\fi
#2}}
\providecommand{\BIBdecl}{\relax}
\BIBdecl

\bibitem{wang2017attribute}
J.~Wang, X.~Zhu, S.~Gong, and W.~Li, ``Attribute recognition by joint recurrent
  learning of context and correlation.''

\bibitem{japuria2017casnsc}
N.~Japuria, G.~Habibi, and J.~P. How, ``Casnsc: A context-based approach for
  accurate pedestrian motion prediction at intersections,'' 2017.

\bibitem{li2016object}
B.~Li, T.~Wu, S.~Shao, L.~Zhang, and R.~Chu, ``Object detection via aspect
  ratio and context aware region-based convolutional networks,'' \emph{arXiv
  preprint arXiv:1612.00534}, 2016.

\bibitem{liu2017face}
S.~Liu, Y.~Sun, D.~Zhu, R.~Bao, W.~Wang, X.~Shu, and S.~Yan, ``Face aging with
  contextual generative adversarial nets,'' in \emph{Proceedings of the 2017
  ACM on Multimedia Conference}.\hskip 1em plus 0.5em minus 0.4em\relax ACM,
  2017, pp. 82--90.

\bibitem{matei2017context}
O.~Matei, T.~Rusu, A.~Bozga, P.~Pop-Sitar, and C.~Anton, ``Context-aware data
  mining: Embedding external data sources in a machine learning process,'' in
  \emph{International Conference on Hybrid Artificial Intelligence
  Systems}.\hskip 1em plus 0.5em minus 0.4em\relax Springer, 2017, pp.
  415--426.

\bibitem{gao2017detecting}
L.~Gao and R.~Huang, ``Detecting online hate speech using context aware
  models,'' \emph{arXiv preprint arXiv:1710.07395}, 2017.

\bibitem{liu2016multi}
R.~Liu, ``Multi-contextual representation and learning with applications in
  materials knowledge discovery,'' Ph.D. dissertation, Northwestern University,
  2016.

\bibitem{mezhoudi2013user}
N.~Mezhoudi, ``User interface adaptation based on user feedback and machine
  learning,'' in \emph{Proceedings of the companion publication of the 2013
  international conference on Intelligent user interfaces companion}.\hskip 1em
  plus 0.5em minus 0.4em\relax ACM, 2013, pp. 25--28.

\bibitem{kabir2015machine}
M.~H. Kabir, M.~R. Hoque, H.~Seo, and S.-H. Yang, ``Machine learning based
  adaptive context-aware system for smart home environment,''
  \emph{International Journal of Smart Home}, vol.~9, no.~11, pp. 55--62, 2015.

\bibitem{wang2010camf}
A.~I. Wang, Q.~K. Ahmad \emph{et~al.}, ``Camf-context-aware machine learning
  framework for android,'' in \emph{Proceedings of the International Conference
  on Software Engineering and Applications (SEA 2010), CA, USA}, 2010.

\bibitem{zhou2012context}
S.~Zhou, C.-H. Chu, Z.~Yu, and J.~Kim, ``A context-aware reminder system for
  elders based on fuzzy linguistic approach,'' \emph{Expert Systems with
  Applications}, vol.~39, no.~10, pp. 9411--9419, 2012.

\bibitem{kwapisz2011activity}
J.~R. Kwapisz, G.~M. Weiss, and S.~A. Moore, ``Activity recognition using cell
  phone accelerometers,'' \emph{ACM SigKDD Explorations Newsletter}, vol.~12,
  no.~2, pp. 74--82, 2011.

\bibitem{widmer1996recognition}
G.~Widmer, ``Recognition and exploitation of contextual clues via incremental
  meta-learning (extended version).''

\bibitem{widmer1996learning}
G.~Widmer and M.~Kubat, ``Learning in the presence of concept drift and hidden
  contexts,'' \emph{Machine learning}, vol.~23, no.~1, pp. 69--101, 1996.

\bibitem{dong2018threaded}
Y.~Dong and N.~Japkowicz, ``Threaded ensembles of autoencoders for stream
  learning,'' \emph{Computational Intelligence}, vol.~34, no.~1, pp. 261--281,
  2018.

\bibitem{zhou2012online}
G.~Zhou, K.~Sohn, and H.~Lee, ``Online incremental feature learning with
  denoising autoencoders,'' in \emph{Artificial intelligence and statistics},
  2012, pp. 1453--1461.

\bibitem{yan2016correcting}
K.~Yan and D.~Zhang, ``Correcting instrumental variation and time-varying
  drift: a transfer learning approach with autoencoders,'' \emph{IEEE
  Transactions on Instrumentation and Measurement}, vol.~65, no.~9, pp.
  2012--2022, 2016.

\bibitem{budiman2016adaptive}
A.~Budiman, M.~I. Fanany, and C.~Basaruddin, ``Adaptive convolutional elm for
  concept drift handling in online stream data,'' \emph{arXiv preprint
  arXiv:1610.02348}, 2016.

\bibitem{widmer1993effective}
G.~Widmer and M.~Kubat, ``Effective learning in dynamic environments by
  explicit context tracking,'' in \emph{European Conference on Machine
  Learning}.\hskip 1em plus 0.5em minus 0.4em\relax Springer, 1993, pp.
  227--243.

\bibitem{lecun1998mnist}
Y.~LeCun, ``The mnist database of handwritten digits,'' \emph{http://yann.
  lecun. com/exdb/mnist/}, 1998.

\bibitem{coraddu2016machine}
A.~Coraddu, L.~Oneto, A.~Ghio, S.~Savio, D.~Anguita, and M.~Figari, ``Machine
  learning approaches for improving condition-based maintenance of naval
  propulsion plants,'' \emph{Proceedings of the Institution of Mechanical
  Engineers, Part M: Journal of Engineering for the Maritime Environment}, vol.
  230, no.~1, pp. 136--153, 2016.

\bibitem{seewald2005digits}
A.~K. Seewald, ``Digits-a dataset for handwritten digit recognition.''

\bibitem{vincent2010stacked}
P.~Vincent, H.~Larochelle, I.~Lajoie, Y.~Bengio, and P.-A. Manzagol, ``Stacked
  denoising autoencoders: Learning useful representations in a deep network
  with a local denoising criterion,'' \emph{Journal of Machine Learning
  Research}, vol.~11, no. Dec, pp. 3371--3408, 2010.

\bibitem{hinton2006reducing}
G.~E. Hinton and R.~R. Salakhutdinov, ``Reducing the dimensionality of data
  with neural networks,'' \emph{science}, vol. 313, no. 5786, pp. 504--507,
  2006.

\bibitem{sakurada2014anomaly}
M.~Sakurada and T.~Yairi, ``Anomaly detection using autoencoders with nonlinear
  dimensionality reduction,'' in \emph{Proceedings of the MLSDA 2014 2nd
  Workshop on Machine Learning for Sensory Data Analysis}.\hskip 1em plus 0.5em
  minus 0.4em\relax ACM, 2014, p.~4.

\bibitem{martinelli2004electric}
M.~Martinelli, E.~Tronci, G.~Dipoppa, and C.~Balducelli, ``Electric power
  system anomaly detection using neural networks,'' in \emph{Knowledge-Based
  Intelligent Information and Engineering Systems}.\hskip 1em plus 0.5em minus
  0.4em\relax Springer, 2004, pp. 1242--1248.

\bibitem{nolle2016unsupervised}
T.~Nolle, A.~Seeliger, and M.~M{\"u}hlh{\"a}user, ``Unsupervised anomaly
  detection in noisy business process event logs using denoising
  autoencoders,'' in \emph{International Conference on Discovery
  Science}.\hskip 1em plus 0.5em minus 0.4em\relax Springer, 2016, pp.
  442--456.

\end{thebibliography}

\end{document}